\documentclass[lettersize,journal]{IEEEtran}

\IEEEoverridecommandlockouts                      %\overrideIEEEmargins                               
\title{\LARGE \bf
TONAV: Task-Oriented Navigation and Action-Velocity Chunk Learning for Articulated Object Quadrupedal Mobile Manipulation}

\author{Haoran Lin$^{1,2,*}$, Mingyu Yang$^{1,2,*}$, Pengfei Qi$^{3}$, Kehan Chen$^{1}$, Qiang Diao$^{1}$,\\Liangji Zeng$^{1}$, Wenrui Chen$^{1,2}$, Yaonan Wang$^{1,2}$, and Kailun Yang$^{1,2,3,}$\textsuperscript{\textdagger}%
\thanks{$^{*}$Equal contribution.}%
\thanks{\textsuperscript{\textdagger}Corresponding author.}%
\thanks{This work was supported in part by the National Natural Science Foundation of China (Grant No. 62473139), in part by the Hunan Provincial Research and Development Project (Grant No. 2025QK3019), and in part by the State Key Laboratory of Autonomous Intelligent Unmanned Systems (the opening project number ZZKF2025-2-10).}
\thanks{$^{1}$The authors are with the School of Artificial Intelligence and Robotics, Hunan University, China.}%
\thanks{$^{2}$The authors are also with the National Engineering Research Center of Robot Visual Perception and Control Technology, Hunan University, China.}%
\thanks{$^{3}$The authors are with the College of Semiconductors (College of Integrated Circuits), Hunan University, Changsha, China.}%
}

\usepackage{placeins}
\usepackage{bbm}
\usepackage{etoolbox}
\usepackage{multicol}
\usepackage{graphics} % for pdf, bitmapped graphics files
\usepackage{times} % assumes a new font selection scheme installed
\usepackage{amsmath} % assumes amsmath package installed
\usepackage{amssymb}  % assumes amsmath package installed
\usepackage{svg}
\usepackage{mathrsfs}
\usepackage{amsfonts}
\usepackage{wrapfig}
\usepackage{amsmath}

\usepackage{kantlipsum}
\usepackage{subcaption}
\usepackage{marvosym}

\usepackage{graphicx}
\usepackage{float}
\usepackage{ctable}
\usepackage{cuted}
\usepackage{colortbl}
\usepackage{multirow}
\usepackage{wrapfig}
\usepackage[misc,geometry]{ifsym}
\usepackage{algorithm}
\usepackage{algpseudocode}

\usepackage{xcolor}   % For color
\usepackage{pifont}    % For check and cross symbols

\definecolor{rblue}{rgb}{0,0.5,1}
\definecolor{awesome}{rgb}{1.0, 0.13, 0.32}
\definecolor{hollywoodcerise}{rgb}{0.96, 0.0, 0.63}
\definecolor{lasallegreen}{rgb}{0.03, 0.47, 0.19}
\definecolor{hanpurple}{rgb}{0.32, 0.09, 0.98}
\definecolor{green(pigment)}{rgb}{0.0, 0.65, 0.31}

\usepackage[pagebackref=false,breaklinks=true,colorlinks,bookmarks=false]{hyperref}
\hypersetup{colorlinks=true,linkcolor={red},citecolor={hanpurple},urlcolor={magenta}}

\usepackage{bm}
\usepackage{caption}
\usepackage{setspace}

\usepackage{enumitem}
\usepackage{threeparttable}
\newcommand{\TODO}[1][]{\textcolor{red}{\bf [TODO]}}
\setlist[enumerate,1]{itemsep=3pt}

\definecolor{formalgreen}{rgb}{0.1, 0.7, 0.1}  % A darker, more formal green
\definecolor{formalred}{rgb}{0.9, 0.2, 0.2}  % A darker, more formal green

\definecolor{rblue}{rgb}{0,0.5,1}

\newcommand{\change}[1]{{\textcolor{black}{#1}}}

\usepackage{xcolor}

\def\LINheader{\hbox{}\scriptsize Lin~\textit{et al}.: TONAV: Task-Oriented Navigation and Action-Velocity Chunk Learning for Articulated Object Quadrupedal Mobile Manipulation \hfil \thepage}

\def\ps@headings{
  \def\@oddhead{\ifcase\value{page}
    \or \RALheader % page 1
    \or \RALheader % page 2
    \or \LINheader % page 3
    \or \RALheader % page 4
    \or \LINheader % page 5
    \or \RALheader % page 6
    \or \LINheader % page 7
    \or \RALheader % page 8
    \else \LINheader % 其他页默认 Lin，或你想改
  \fi}
  \def\@evenhead{\@oddhead}
  \def\@oddfoot{}\def\@evenfoot{}
}
\makeatother

\begin{document}

\maketitle
\thispagestyle{IEEEtitlepagestyle} 
\pagestyle{headings}               

\begin{abstract}
Quadruped mobile manipulation requires two tightly coupled capabilities: reaching manipulation-ready configurations and maintaining stable contact throughout articulated-object interaction. However, existing methods often terminate navigation near the target, leaving a gap between reachability and manipulation readiness, while tracking lag, motion jitter, and contact instability limit continuous interaction. To address these challenges, we present TONAV, a unified framework integrating task-oriented navigation with action-velocity chunk learning. First, we introduce a position-velocity-coupled teleoperation framework that explicitly captures motion dynamics to improve master-follower consistency and collect smooth, temporally consistent demonstrations. Next, task-oriented navigation leverages vision-language reasoning to decompose high-level instructions into executable subgoals and adaptively refine the robot base toward a manipulation-ready configuration. Finally, action-velocity chunk learning jointly models joint positions and their temporal transitions under velocity supervision, enabling smooth and stable sustained-contact manipulation. Real-world experiments across diverse articulated-object tasks demonstrate that TONAV achieves higher success rates in both task-oriented navigation and complete mobile manipulation, mitigating the navigation-manipulation gap and improving continuous-contact interaction. The project page is at~\href{https://haochen611.github.io/TONAV}{https://haochen611.github.io/TONAV}.
\end{abstract}

\section{Introduction}
Mobile manipulation requires robots to navigate to task-relevant objects or regions and subsequently manipulate them~\cite{DBLP:conf/iros/YangKHKWE24,DBLP:conf/cvpr/WuZXXWY25,DBLP:journals/corr/abs-2605-03846}, which is a core capability for household service robots~\cite{DBLP:conf/icra/XiaoJHGLYH25,DBLP:conf/icra/AbbatematteoRTA24}.
In such tasks, navigation should terminate at a manipulation-ready base pose---a configuration that is not only close to the target but also compatible with the intended interaction.
{However, existing object-navigation systems are primarily designed to reach the vicinity of a target~\cite{DBLP:conf/rss/ChangGKYSMS0GBM24,DBLP:conf/icra/WeiWYMCZCYWCWL26,DBLP:journals/corr/abs-2512-08186,DBLP:conf/icra/HiroseGSL26}, \textit{e.g.}, within 1--2 meters, which is substantially coarser than required for manipulation.
This granularity mismatch can leave the robot poorly aligned, out of reach, or obstructed during subsequent interaction~\cite{DBLP:conf/iccv/ZhangGWLWWZDL25}.
This leads to the first core challenge: task-oriented navigation, where the robot must go beyond target proximity and progressively adjust its base configuration according to task-specific spatial and kinematic constraints to reach a manipulation-ready pose~\cite{NEURIPS2025_23bff092}.

Existing approaches address this problem by manually specifying task-specific base poses~\cite{DBLP:conf/icra/QiuSPSYLJJYZW25}, learning object affordances~\cite{Zhang2026UniLMNav,Lin2026AffordanceGuided}, estimating manipulation reachability~\cite{DBLP:conf/iccv/ZhangGWLWWZDL25}, or optimizing geometric and kinematic feasibility~\cite{Liu2025VisualWholeBodyControl}.
These methods either rely on task-specific annotations, extensive training data, or predefined geometric criteria, which limits their flexibility and generalization to unseen tasks and environments.
Recent vision foundation models~\cite{Kirillov2023SegmentAnything,Oquab2024DINOv2,Radford2021CLIP} and multimodal large language models~\cite{Hurst2024GPT4o,OpenAI2025O3O4Mini} provide new opportunities for open-vocabulary perception and task-level visual reasoning.
Recent methods have therefore explored reasoning about last-mile base poses from object-level cues~\cite{Wu2025MoTo}.
Nevertheless, directly predicting a final base pose from the target and nearby objects provides limited reasoning over navigation structure and fine-grained spatial constraints, particularly in narrow household environments.
For tasks such as ``close the cabinet'', the robot must instead interpret the high-level instruction, identify task-relevant landmarks and free space, and execute a sequence of navigation behaviors before progressively refining its final approach.

\begin{figure}[!t]
    \centering
    \includegraphics[width=1.0\linewidth]{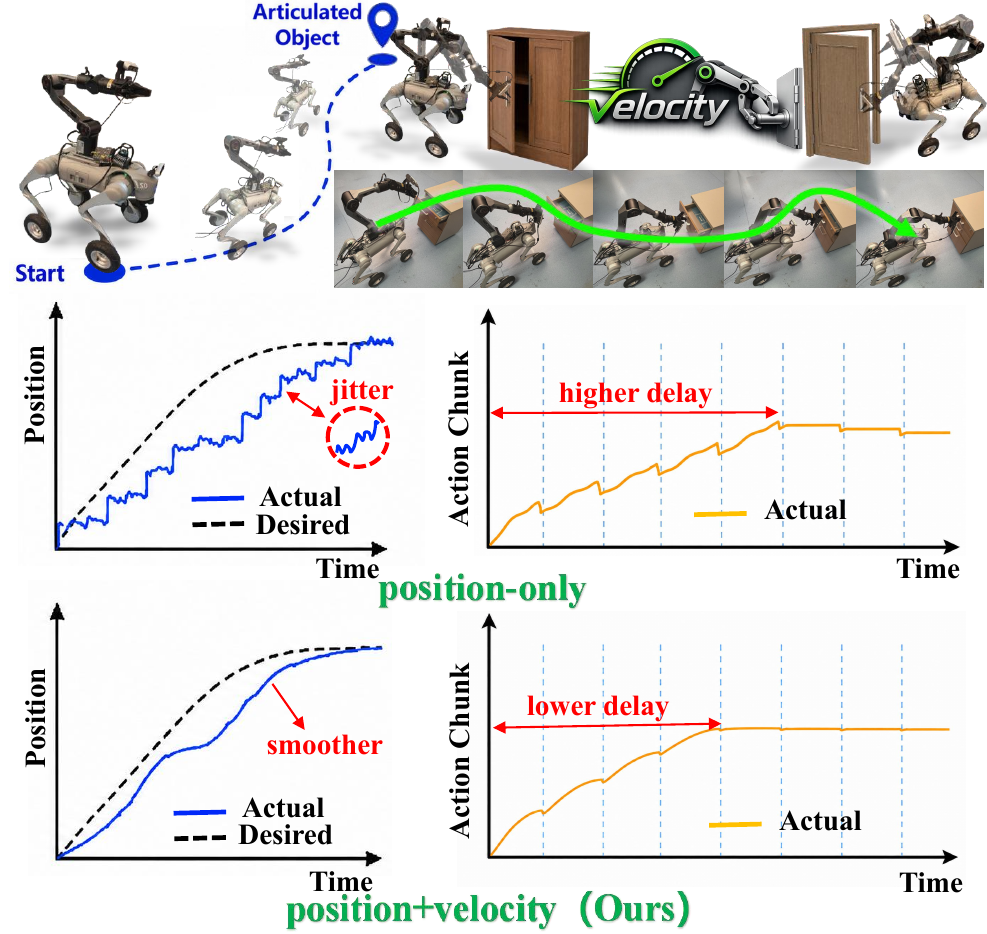}
    \vskip-1ex
    \caption{Illustration of articulated-object quadrupedal mobile manipulation, with progressive target approaching followed by position-velocity-based sustained-contact manipulation for smoother and more stable interaction.}
    \label{fig:figure1}
    \vskip-3ex
\end{figure}

Reaching a manipulation-ready configuration, however, is only part of the problem. A second coupled challenge is maintaining continuous and stable interaction after contact is established.
Existing methods commonly characterize manipulation-ready poses using static criteria such as reachability~\cite{Zhang2026UniLMNav}, orientation~\cite{Yang2025MobiPi}, or spatial clearance~\cite{DBLP:conf/iccv/ZhangGWLWWZDL25}.
While these criteria are useful for initiating manipulation, they do not capture the evolving motion state required during sustained interaction.
This issue is especially important for articulated-object manipulation~\cite{Zhao2025TacMan,Pi2026CoDA,Engelbracht2026HOI,Wu2026AOMGen,Wu2026ArticulatedParts}, such as closing drawers, operating cabinets, and manipulating toilet lids, where the robot must establish contact and continuously follow constrained object trajectories.
Successful execution therefore depends not only on where the robot and end-effector should move, but also on how their motion evolves throughout the interaction.

\begin{figure*}[!t] \centering
    \includegraphics[width=0.98\linewidth]{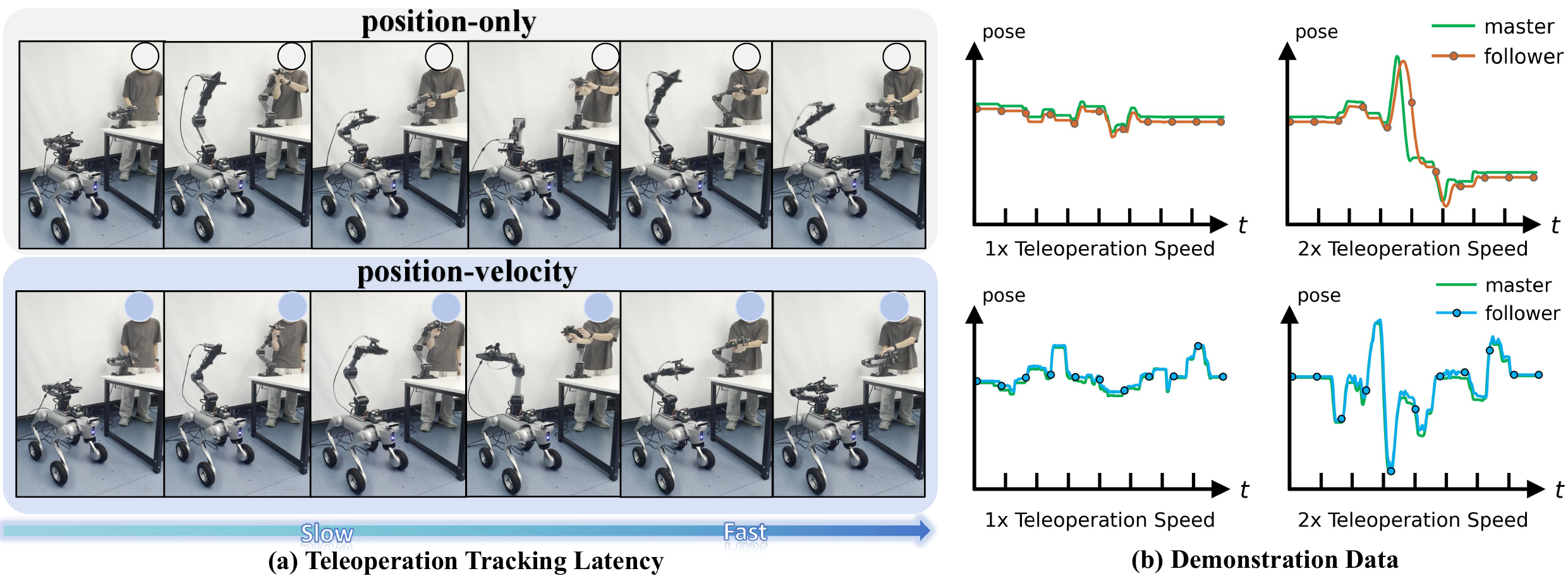}
    \vskip-1ex
    \caption{\change{Comparison of position-only and position-velocity-coupled teleoperation. (a) Teleoperation at increasing speeds, showing reduced tracking lag with position-velocity coupling. (b) Master-follower trajectories at $1\times$ and $2\times$ speeds, showing lower latency and better temporal consistency.
}} \label{fig:2}
%
%\vskip-1ex
\end{figure*}

Learning-based manipulation policies~\cite{Zhao2023ACT,Chi2023DiffusionPolicy} provide an effective way to learn such contact-rich behaviors from expert demonstrations.
Action-chunking policies, in particular, predict sequences of future joint configurations and have demonstrated strong performance in fine-grained manipulation.
Yet, most existing formulations represent actions primarily through joint positions, without explicitly modeling the corresponding motion velocities.
Position specifies where the robot should move, but does not fully characterize how the motion should evolve between successive configurations.
Consequently, position-only action chunks can exhibit delayed responses, discontinuous motion, and unstable contact, especially during sustained interaction, chunk transitions, or variations in the initial robot configuration.

These two coupled challenges motivate our framework: reaching a manipulation-ready configuration from high-level task instructions and maintaining smooth, stable interaction throughout manipulation.
To address the aforementioned challenges, we propose \textbf{TONAV}, a unified framework that integrates task-oriented navigation with action-velocity chunk learning for articulated-object quadrupedal mobile manipulation.
The navigation stage interprets high-level instructions into executable navigation subgoals and progressively refines the robot base through last-mile navigation toward a manipulation-ready configuration.
For manipulation, a position-velocity-coupled teleoperation scheme collects temporally consistent demonstrations, while action-velocity chunk learning jointly models future joint positions and their motion velocities.
Together, these components establish a coherent transition from task-driven target approach to smooth and stable continuous-contact manipulation.

Real-world experiments on three articulated-object mobile manipulation tasks show that TONAV achieves an overall navigation SR of $60.00\%$, with $80.00\%$ success across all three manipulation tasks under shared navigation outcomes. Ablation studies and trajectory analysis further validate the contributions of PP-CoT, position-velocity-coupled teleoperation, and action-velocity chunk learning to manipulation-ready navigation and stable sustained-contact manipulation.

Our main contributions are summarized as follows:
\begin{itemize}
\item We propose \textbf{TONAV}, a hybrid-policy framework for articulated-object mobile manipulation. It leverages Perception-Planning Chain-of-Thought (PP-CoT) reasoning to decompose high-level instructions into sequential navigation subgoals, followed by an Adaptive Approach Module (AAM) that progressively refines the robot base toward a manipulation-ready configuration.

\item We develop a position-velocity-coupled teleoperation scheme that integrates joint-velocity references into impedance-based tracking, reducing master-follower motion mismatch and response latency while providing smoother and more temporally consistent expert demonstrations.

\item We propose an action-velocity chunk learning method that jointly predicts future joint-position and joint-velocity sequences, extending position-only action chunks from modeling \emph{where to move} to also capturing \emph{how to move}. 
This explicit motion modeling enables smoother, more responsive, and more stable sustained-contact manipulation of articulated objects.
\end{itemize}

\section{Related Work}

\subsection{Embodied Navigation}
Existing navigation methods can be broadly divided into two levels: coarse-grained target-directed navigation~\cite{DBLP:conf/rss/ChangGKYSMS0GBM24, DBLP:conf/icra/WeiWYMCZCYWCWL26, DBLP:journals/corr/abs-2512-08186, DBLP:conf/icra/HiroseGSL26} and fine-grained last-mile navigation~\cite{DBLP:conf/iccv/ZhangGWLWWZDL25, Lin2026AffordanceGuided, Wu2025MoTo}. Target-directed navigation aims to guide the robot over relatively long distances toward the vicinity of a target. Recent vision-language navigation~\cite{DBLP:conf/icra/WeiWYMCZCYWCWL26, DBLP:journals/corr/abs-2512-08186, DBLP:conf/icra/HiroseGSL26} and open-vocabulary navigation methods~\cite{Zhang2026UniLMNav} further enable legged robots to follow semantic goals or language instructions in unseen environments. For last-mile navigation, some approaches learn navigation policies through imitation learning~\cite{Lee2025LastMeter} or reinforcement learning~\cite{Qin2025Active}; MoMa-Kitchen~\cite{DBLP:conf/iccv/ZhangGWLWWZDL25} directly learns distributions over manipulation-ready base poses from large-scale supervised data; another line of work further conditions base-pose preferences on downstream manipulation policies~\cite{Yang2025MobiPi,Chai2025N2M}. However, existing methods typically focus either on \emph{how to reach the target vicinity} or on \emph{how to refine the base pose after reaching the vicinity}, lacking a unified closed-loop process from long-range target-directed navigation to fine-grained last-mile adjustment. As a result, they do not fully address task-oriented navigation tailored to downstream manipulation.

\subsection{Mobile Manipulation}
Existing approaches to mobile manipulation can be broadly divided into end-to-end methods and modular methods. End-to-end methods~\cite{Brohan2023RT1,Fu2025MobileALOHA,Yan2025M2Diffuser} learn a joint policy for base and arm actions from large-scale demonstrations, but scaling them to open-vocabulary and long-horizon tasks often incurs substantial data collection costs. Modular approaches~\cite{Yenamandra2023HomeRobot,Liu2024OKRobot,Rosen2023NavigationAbstractions} decouple navigation from manipulation, thereby improving data efficiency and system scalability. These methods typically rely on heuristic strategies to determine manipulation-ready base poses, including pose scoring~\cite{Liu2024OKRobot}, sampling-based region estimation~\cite{Rosen2023NavigationAbstractions}, feasibility predicates~\cite{Wang2025InstructionAugmented}, trajectory back-projection~\cite{DBLP:conf/cvpr/WuZXXWY25}, and motion planning~\cite{Quartey2025VerifiablyFollowing}. Although these strategies help bridge navigation and manipulation, their reliance on predefined heuristics can limit their adaptability and flexibility in articulated-object manipulation tasks~\cite{Zhao2025TacMan,Pi2026CoDA,Engelbracht2026HOI,Wu2026AOMGen,Wu2026ArticulatedParts} that require precise base positioning and continuous-contact constraints.

\begin{figure}[!t]
    \centering
    \includegraphics[width=\linewidth]{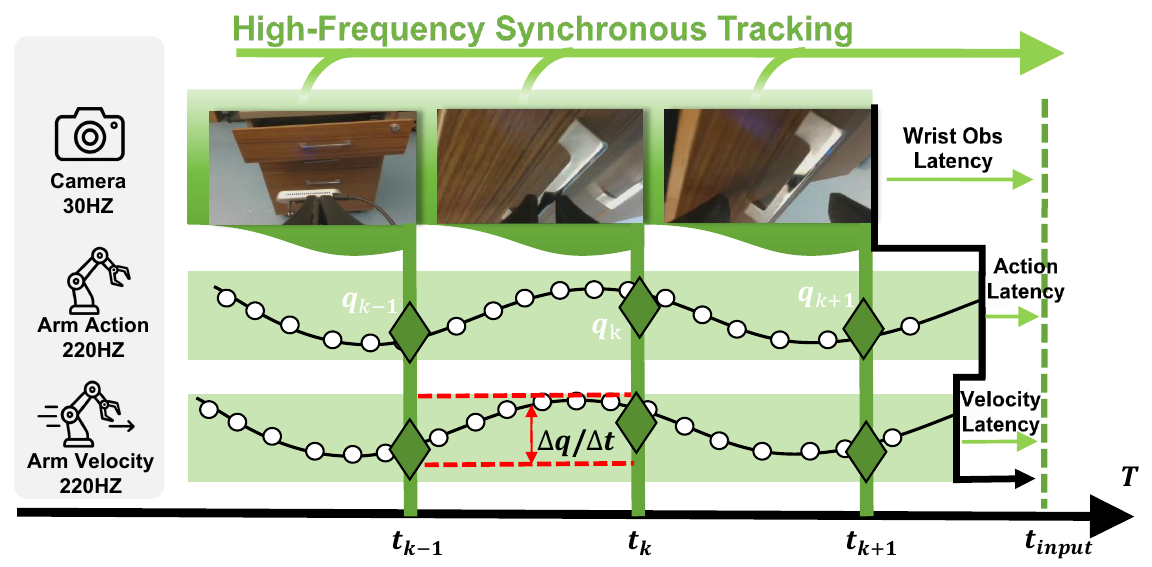}
    \vskip-1ex
    \caption{Position-velocity-coupled teleoperation with $220\,\mathrm{Hz}$ position-velocity tracking, $30\,\mathrm{Hz}$ camera observations, and timestamp-based alignment across latency-varying modalities.}
    \vskip-1ex
    \label{fig:3}
\end{figure}

\section{Method}
\begin{figure*}[!t] \centering
    \includegraphics[width=0.98 \linewidth]{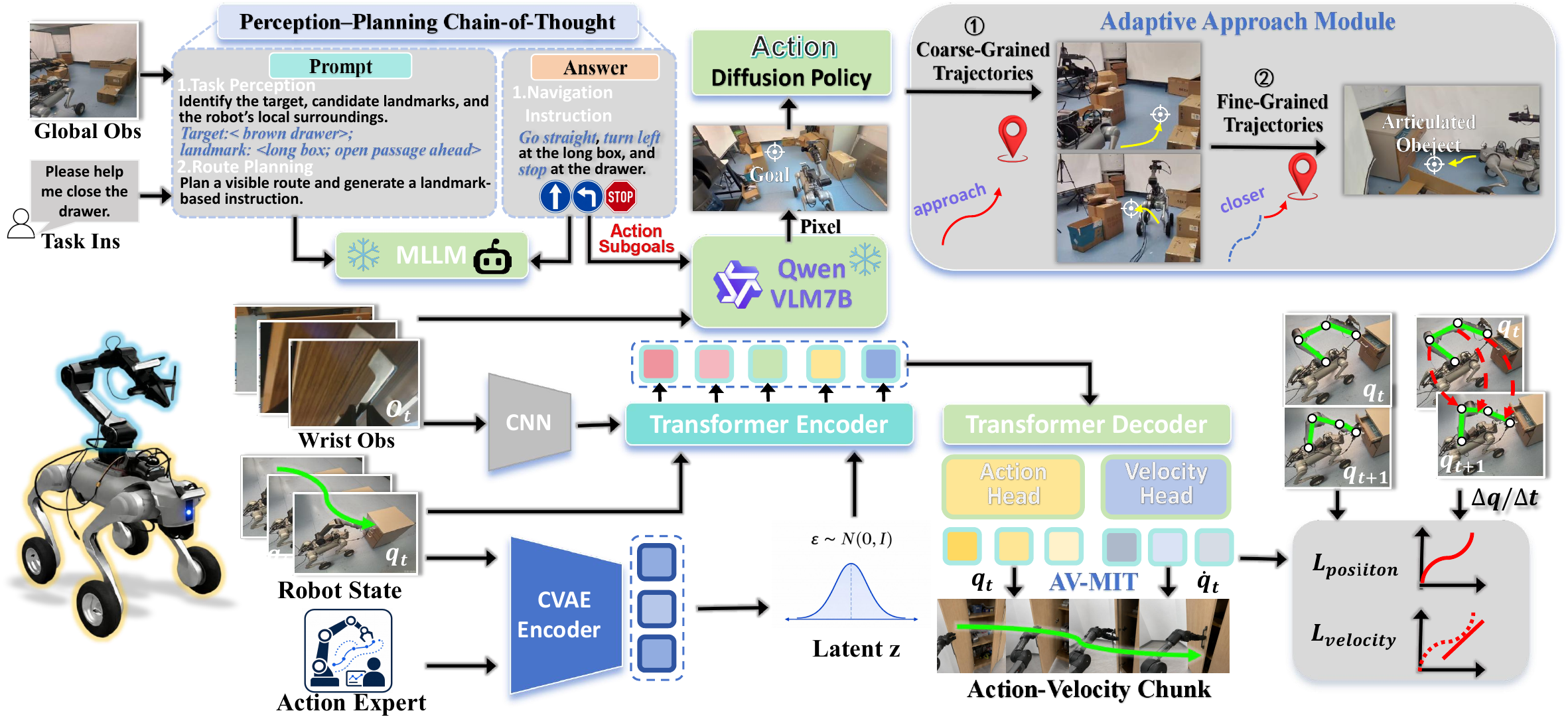}
    \vskip-1ex
    \caption{\change{Overview of TONAV with an upper navigation branch and a lower manipulation branch. The upper branch combines hierarchical task reasoning and adaptive last-mile navigation to reach a manipulation-ready configuration, while the lower branch learns action-velocity chunks from motion-consistent demonstrations for smooth and stable sustained-contact manipulation.}} 
    \label{fig:pipeline}
    \vskip-1ex
    \label{fig:4}
\end{figure*}

\begin{figure}[!t]
    \centering
    \includegraphics[width=\linewidth]{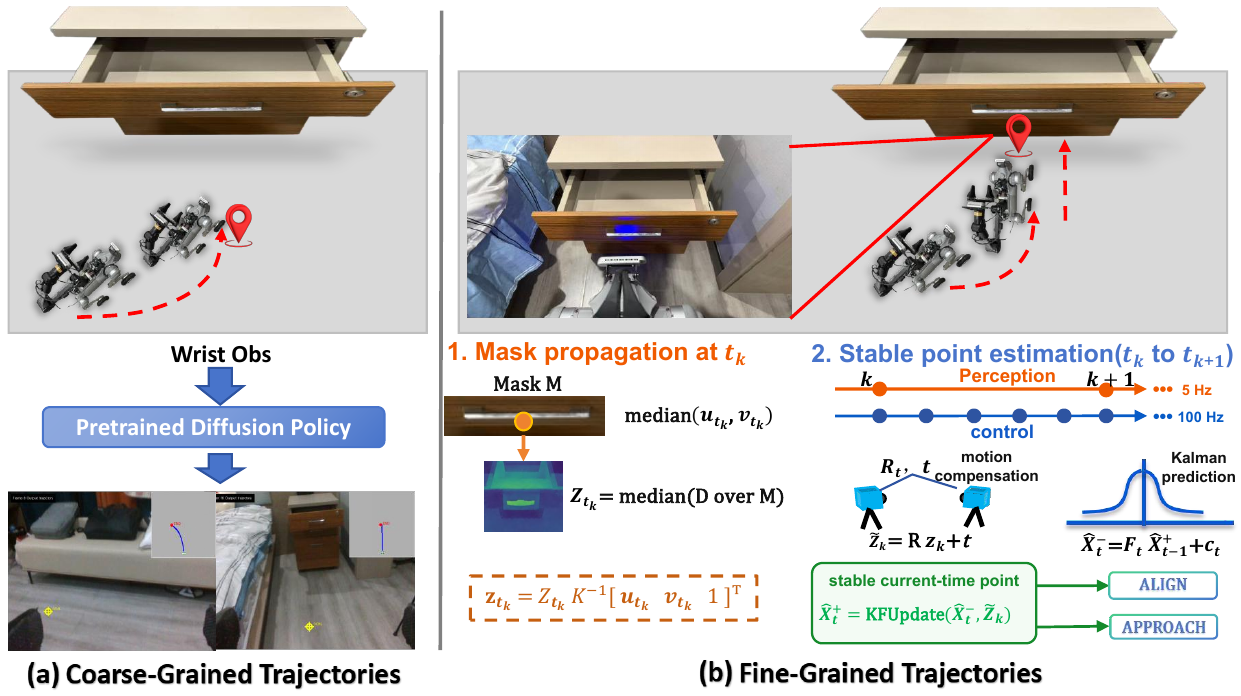}
    \vskip-1ex
    \caption{Illustration of the Adaptive Approach Module (AAM). 
    AAM performs coarse approaching with a diffusion policy and fine target refinement with visual feedback.}
    \vskip-3ex
    \label{fig:5}
\end{figure}

In this work, we enable quadrupedal robots to navigate toward manipulation-ready base poses and perform stable continuous-contact interaction with articulated objects. Given language instructions, visual observations, and proprioceptive states, TONAV unifies task-oriented navigation and contact-rich manipulation, bridging coarse target-directed navigation with fine-grained manipulation through explicit joint position-velocity modeling.
An overview is shown in Fig.~\ref{fig:4}.

\textbf{Method Overview.}
First, we introduce an Action-Velocity Teleoperation Framework (AVTF) (see Sec.~\ref{subsec:teleoperation}) that incorporates joint-velocity references into impedance-based master-follower tracking to collect smoother and more temporally consistent demonstrations. 
Next, a Perception-Planning Chain-of-Thought
(PP-CoT) (see Sec.~\ref{subsec:COT}) decomposes high-level instructions into executable navigation subgoals, while an Adaptive Approach Module (AAM) (see Sec.~\ref{sec:AAM}) progressively refines the base pose through last-mile navigation to reach a manipulation-ready configuration. 
Finally, an Action-Velocity Chunk Learning policy (see Sec.~\ref{subsec:AVchunk}) jointly predicts future joint-position and joint-velocity sequences, enabling smooth, responsive, and stable continuous-contact manipulation.

\begin{figure*}[t]
    \centering
    \includegraphics[width=0.98\linewidth]{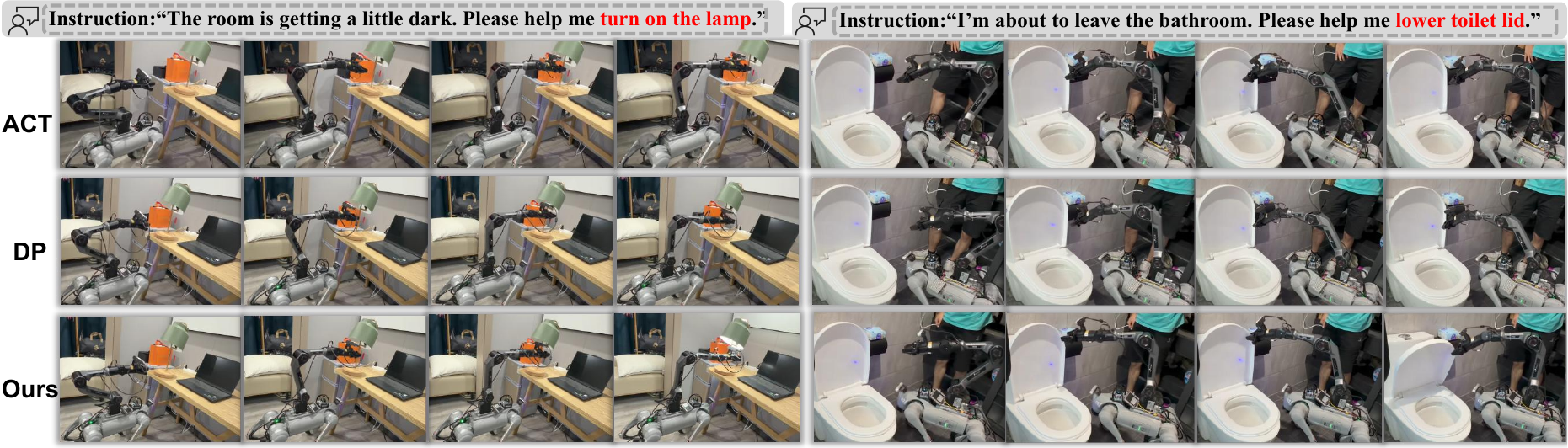}
    \vskip-1ex
    \captionsetup{justification=raggedright,singlelinecheck=false}
    \caption{Qualitative comparison on the two tasks. 
    Ours leverages action–velocity demonstrations and chunk learning to achieve smoother motion and more stable sustained-contact manipulation than ACT~\cite{Zhao2023ACT} and DP~\cite{Chi2023DiffusionPolicy}.}
    \label{fig:6}
    \vskip-2ex
\end{figure*}

\subsection{Action-Velocity Teleoperation Framework}
\label{subsec:teleoperation}
Conventional teleoperation typically transmits only joint-position references from the master arm to the follower arm.
Although position references specify where the follower should move, they do not explicitly encode the corresponding motion direction and rate. This limitation is particularly severe for low-cost manipulators with limited compliant-control capability, where position-only tracking can amplify response lag, motion discontinuities, and contact instability. 
Such effects are especially detrimental to articulated-object manipulation, which requires sustained contact along constrained trajectories. 
To improve demonstration quality, we introduce an Action-Velocity Teleoperation Framework (AVTF), which couples joint-position commands with their corresponding joint-velocity references for follower-arm control. 
As illustrated in Fig.~\ref{fig:2}, position-velocity coupling reduces tracking lag during teleoperation and produces more temporally consistent master--follower trajectories.

At control step $t$, the joint configuration of the master arm is denoted by
\begin{equation}
\mathbf{q}_{t}^{m}
=
\left[
q_{t,1}^{m},
\ldots,
q_{t,N}^{m}
\right]^{\top}
\in \mathbb{R}^{N},
\end{equation}
where $N$ is the number of arm joints. The corresponding joint velocity is estimated from consecutive master configurations using a backward finite difference:
\begin{equation}
\hat{\dot{\mathbf{q}}}_{t}^{m}
=
\frac{
\mathbf{q}_{t}^{m}
-
\mathbf{q}_{t-1}^{m}
}{
\Delta t_{t}
},
\end{equation}
where $\Delta t_t$ denotes the actual interval between two consecutive measurements. To suppress excessively large velocity references caused by measurement noise or timing fluctuations, each estimated joint velocity is bounded as
\begin{equation}
\dot{q}_{t,i}^{m,\mathrm{ref}}
=
\operatorname{clip}
\left(
\hat{\dot{q}}_{t,i}^{m},
-\dot{q}_{i}^{\max},
\dot{q}_{i}^{\max}
\right),
\qquad i=1,\ldots,N,
\end{equation}
where $\dot{q}_{i}^{\max}$ denotes the velocity limit of the $i$-th joint. The bounded velocity references form
$\dot{\mathbf{q}}_{t}^{m,\mathrm{ref}}\in\mathbb{R}^{N}$.

The teleoperation command is then represented as a coupled action--velocity reference:
\begin{equation}
\mathbf{a}_{t}^{\mathrm{tele}}
=
\begin{bmatrix}
\mathbf{q}_{t}^{m}\\
\dot{\mathbf{q}}_{t}^{m,\mathrm{ref}}
\end{bmatrix}
\in\mathbb{R}^{2N}.
\end{equation}
As shown in Fig.~\ref{fig:3}, the follower arm tracks this reference through an MIT-style impedance controller~\cite{wensing2017proprioceptive,jenelten2019dynamic}:
\begin{equation}
\boldsymbol{\tau}_{t}
=
\mathbf{K}_{p}
\left(
\mathbf{q}_{t}^{m}
-
\mathbf{q}_{t}^{f}
\right)
+
\mathbf{K}_{d}
\left(
\dot{\mathbf{q}}_{t}^{m,\mathrm{ref}}
-
\dot{\mathbf{q}}_{t}^{f}
\right)
+
\boldsymbol{\tau}_{\mathrm{ff}},
\end{equation}
where $\mathbf{q}_{t}^{f}$ and $\dot{\mathbf{q}}_{t}^{f}$ are the measured joint position and velocity of the follower arm, respectively. The matrices $\mathbf{K}_{p}$ and $\mathbf{K}_{d}$ denote the joint-position stiffness and velocity damping gains, and $\boldsymbol{\tau}_{\mathrm{ff}}$ is the feedforward torque. 
The position term regulates configuration tracking, while the velocity term conveys the local motion trend of the master arm. Their combination improves follower responsiveness and reduces temporal and trajectory mismatch during continuous interaction.

To prevent visual acquisition and data storage from interrupting robot control, AVTF decouples the teleoperation-control loop from the demonstration-recording loop. The follower controller runs in an independent high-frequency control thread, while visual observations and follower-arm feedback are recorded in a separate recording loop. 
At each recording step \(k\), the demonstration frame is saved as
\begin{equation}
\mathcal{D}_{k}
=
\left\{
\mathbf{o}_{k},
\mathbf{q}_{k}^{f},
\mathbf{q}_{k}^{m},
\dot{\mathbf{q}}_{k}^{m}
\right\},
\end{equation}
where \(\mathbf{o}_{k}\) denotes the visual observation, \(\mathbf{q}_{k}^{f}\) is the follower-arm proprioceptive feedback, \(\mathbf{q}_{k}^{m}\) denotes the expert action provided by the master arm, and \(\dot{\mathbf{q}}_{k}^{m}\) is the corresponding action velocity estimated from consecutive master-arm actions. 
Specifically, \(\mathbf{q}_{k}^{m}\) is converted to joint angles in radians and gripper position in meters, while \(\dot{\mathbf{q}}_{k}^{m}\) represents the corresponding velocity in rad/s and m/s. 
Recording the follower-arm feedback together with the expert actions and their velocities provides temporally consistent demonstrations for subsequent action--velocity chunk learning.

\subsection{Perception-Planning Chain-of-Thought}
\label{subsec:COT}
High-level manipulation instructions do not directly specify navigation targets or executable motions, requiring the robot to jointly understand the scene, ground the target, and plan a feasible route. Performing these steps in a single inference may introduce grounding or planning errors. 
To address this, we propose a Perception-Planning Chain-of-Thought (PP-CoT) that progressively converts high-level instructions into executable navigation sub-instructions through task-conditioned scene perception and perception-guided route planning.

Given a global observation $I$ and task instruction $T$, the model first extracts a task-relevant scene representation $P$ and then generates navigation sub-instructions $Y$:
\begin{align}
P &= f_{\theta}\left(I,T;p_{\mathrm{per}}\right),
\label{eq:ppcot_perception}\\
Y &= f_{\theta}\left(I,T,P;p_{\mathrm{plan}}\right),
\label{eq:ppcot_planning}
\end{align}
where $f_{\theta}$ denotes the MLLM, and $p_{\mathrm{per}}$ and $p_{\mathrm{plan}}$ are the perception and planning prompts, respectively.

\textbf{Task-Conditioned Scene Perception:}
The first stage extracts the task target, candidate landmarks, and traversable regions while filtering task-irrelevant objects. The perception prompt is:
{``Identify the task target, candidate landmarks, and the robot's local surroundings.''}
Visible and distinguishable landmarks near potential route changes are retained as navigation anchors, yielding a compact task-relevant representation $P$.

\textbf{Perception-Guided Route Planning:}
Given $P$ and $I$, the model reasons about robot-centric spatial relationships, taking the image bottom center as the robot position and the image centerline as the forward direction. It plans a visible route toward the target while avoiding obstacles and scene boundaries, and associates directional changes with corresponding landmarks. The planning prompt is:
{``Plan a visible route and generate landmark-based navigation sub-instructions.''}
The resulting route is converted into executable sub-instructions, including moving forward, turning at landmarks, approaching the target, and stopping.

\subsection{Adaptive Approach Module}
\label{sec:AAM}
Conventional object navigation typically terminates near the target, without guaranteeing a manipulation-ready base pose. To bridge this gap, we propose an Adaptive Approach Module (AAM) for target-aligned near-field refinement after upstream navigation.

First, given a language query, AAM grounds the target using an open-vocabulary detector and initializes SAM2 with the detected bounding box. At each visual update $t_k$, AAM computes the coordinate-wise median $(u_{t_k},v_{t_k})$ of the propagated mask and the median valid depth $Z_{t_k}$. These quantities are back-projected to obtain a 3D target observation:
\begin{equation}
\mathbf{z}_{t_k}=Z_{t_k}\mathbf{K}^{-1}
\begin{bmatrix}
u_{t_k}\\
v_{t_k}\\
1
\end{bmatrix},
\label{eq:aam_3d_observation}
\end{equation}
where $\mathbf{K}$ denotes the camera intrinsic matrix. The resulting point represents the target position relative to the onboard camera and provides the visual measurement for subsequent state estimation.

Second, asynchronous SAM2 inference may return an observation after the camera has moved from its capture pose, leading to a spatiotemporal mismatch with the current control state. To compensate for this delay, AAM maintains the target position--velocity state
$\mathbf{x}_t=[\mathbf{p}_t^{\mathsf{T}},\mathbf{v}_t^{\mathsf{T}}]^{\mathsf{T}}$
using an ego-motion-aware Kalman estimator. At the control rate, the state is propagated using constant-velocity dynamics and robot odometry. When an observation captured at $t_k$ becomes available at time $t$, it is transformed from $C_{t_k}$ to the current camera frame $C_t$ and used for measurement correction:
\begin{equation}
\begin{aligned}
\hat{\mathbf{x}}_t^{-}
&=
\mathbf{F}_{t}\hat{\mathbf{x}}_{t-1}^{+}
+\mathbf{c}_{t},\\
\tilde{\mathbf{z}}_t
&=
{}^{C_t}\mathbf{R}_{C_{t_k}}\mathbf{z}_{t_k}
+
{}^{C_t}\mathbf{t}_{C_{t_k}},\\
\hat{\mathbf{x}}_t^{+}
&=
\operatorname{KFUpdate}
\left(
\hat{\mathbf{x}}_t^{-},
\tilde{\mathbf{z}}_t
\right).
\end{aligned}
\label{eq:aam_delayed_kalman}
\end{equation}
Here, $\mathbf{F}_{t}$ and $\mathbf{c}_{t}$ encode the constant-velocity dynamics and odometry-derived camera motion, while
${}^{C_t}\mathbf{R}_{C_{t_k}}$ and
${}^{C_t}\mathbf{t}_{C_{t_k}}$ compensate for the camera motion between image capture and observation arrival. The estimator therefore provides a control-rate target state while incorporating delayed visual updates whenever they become available.

Third, AAM derives the normalized horizontal alignment error and stand-off distance error from the filtered target state:
\begin{equation}
e^u_t=\frac{\hat{u}_t-W/2}{W/2}, \quad
e^d_t=\hat{Z}_t-d^\star ,
\label{eq:aam_control_errors}
\end{equation}
where $\hat{u}_t$ and $\hat{Z}_t$ denote, respectively, the horizontal image coordinate and depth obtained from the filtered target position $\hat{\mathbf{p}}_t$, $W$ is the image width, and $d^\star$ is the desired stand-off distance. Rather than approaching the target as closely as possible, $d^\star$ is selected according to the robot's manipulation workspace so that the end-effector remains within a suitable operating range while avoiding excessive arm extension or potential collision. Based on these errors, a hybrid controller alternates between target-alignment and distance-regulated approach modes: $e^u_t$ regulates angular motion to center the target, while $e^d_t$ regulates forward motion toward the manipulation-ready configuration. A safe stop is triggered whenever perception becomes unreliable or a local safety constraint is violated.

\begin{table}[!t]
\centering
\caption{
Comparison of navigation Success Rates (SR) in the first stage across three articulated-object loco-manipulation tasks, together with the Overall SR.
}
\label{tab:success_rate}
\resizebox{\columnwidth}{!}{
\begin{tabular}{lcccc}
\toprule
& \multicolumn{3}{c}{\textbf{Navigation Success Rates}} 
& \textbf{Overall} \\
\cmidrule(lr){2-4}
\textbf{Method}
& \textbf{Close Drawer} 
& \textbf{Turn On Lamp}
& \textbf{Lower Toilet Lid}
& \textbf{SR}\\
\midrule

StreamVLN~\cite{DBLP:conf/icra/WeiWYMCZCYWCWL26}
& 0/5 & 1/5 & 1/5 & 13.33\% \\

InternNav~\cite{wang2025internvla}
& 2/5 & 1/5 & 1/5 & 26.67\% \\

\textbf{TONAV (Ours)}
& \textbf{4/5}
& \textbf{3/5}
& \textbf{2/5}
& \textbf{60.00\%} \\

\bottomrule
\end{tabular}
}
%\vspace{-10pt}
\label{tab:1}
\vskip-4ex
\end{table}
\begin{figure*}[!t]
    \centering
    \includegraphics[width=0.98\linewidth]{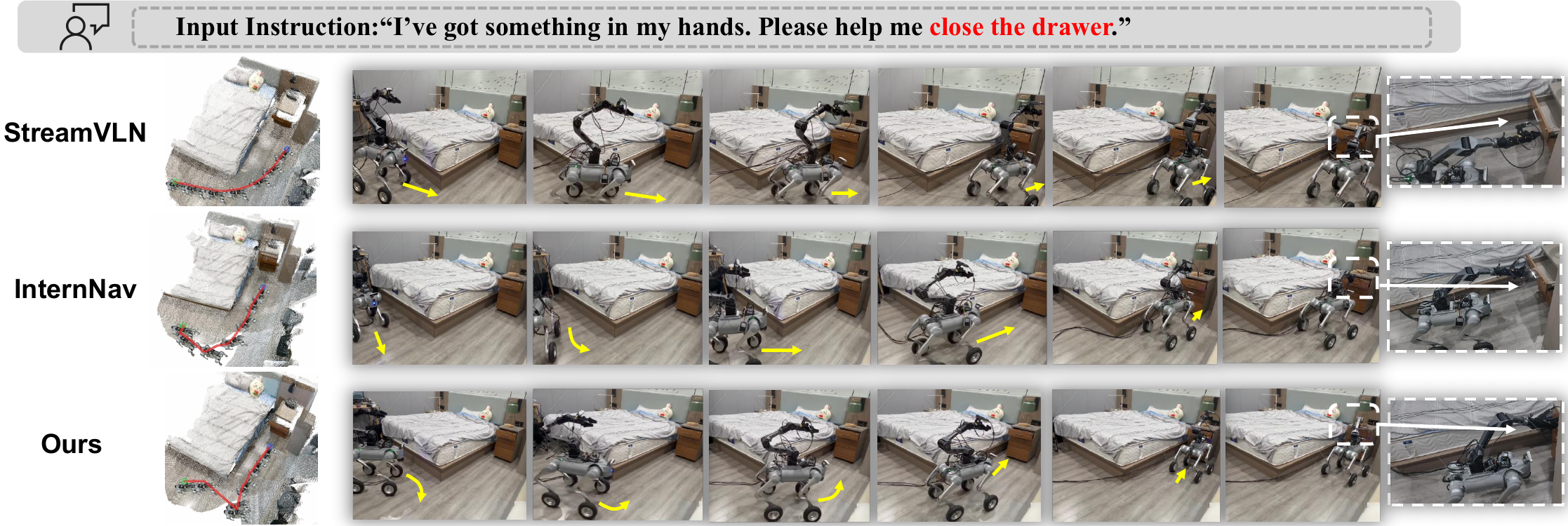}
    \vskip-1ex
    \captionsetup{justification=raggedright,singlelinecheck=false}
    \caption{Qualitative comparison of task-oriented navigation. Ours progressively approaches the target, refines the base pose for manipulation readiness, and closes the drawer within the manipulable workspace using AVTF-based control (see Sec.~\ref{subsec:teleoperation}).}
    \label{fig:7}
    %\vspace{-15pt}
    \vskip-1ex
\end{figure*}

\subsection{Action-Velocity Chunk Learning for Continuous Contact}
\label{subsec:AVchunk}

\textbf{Action-Velocity Chunk Policy:}
Articulated-object mobile manipulation requires manipulation-ready base configurations and stable contact during constrained motion. However, position-only action chunking may cause rollback, oscillation, and contact instability. We therefore propose an action--velocity chunk policy that jointly predicts future joint positions and velocities.

As shown in Fig.~\ref{fig:pipeline}, after task-oriented navigation reaches a manipulation-ready configuration, the policy takes the RGB observation $\mathbf{o}_t$ and proprioceptive state $\mathbf{q}_t$ as inputs. The image and robot state are encoded and fused within an ACT-style CVAE~\cite{Zhao2023ACT}, where the encoder infers a latent variable $\mathbf{z}$ from $\mathbf{q}_t$ and the expert action chunk $\mathbf{q}_{t:t+H}$. The decoder then employs two parallel heads to predict the position chunk $\hat{\mathbf{q}}_{t:t+H}$ and velocity chunk $\hat{\dot{\mathbf{q}}}_{t:t+H}$. The recorded velocity is used only for supervision. A motion-consistency constraint further couples the two predictions by enforcing consistency between the temporal variation of predicted positions and the predicted velocities, thereby promoting smoother motion transitions and more stable sustained-contact manipulation.

\textbf{Loss Functions:}
The policy is trained with position imitation, velocity supervision,
motion consistency, and KL regularization:
\begin{equation}
\mathcal{L}
=
\mathcal{L}_{\mathrm{pos}}
+
\lambda_v \mathcal{L}_{\mathrm{vel}}
+
\lambda_c \mathcal{L}_{\mathrm{con}}
+
\beta \mathcal{L}_{\mathrm{KL}},
\end{equation}
where $\lambda_v$, $\lambda_c$, and $\beta$ weight the corresponding terms.
The position imitation loss is
\begin{equation}
\mathcal{L}_{\mathrm{pos}}
=
\left\|
\hat{\mathbf{q}}_{t:t+H}
-
\mathbf{q}_{t:t+H}
\right\|_1 ,
\end{equation}
where $\mathbf{q}_{t:t+H}$ denotes the reached joint-position sequence recorded
from the follower arm. Using physically executed positions as supervision
reduces the mismatch between master commands and realized robot trajectories.
The velocity supervision is defined as
\begin{equation}
\mathcal{L}_{\mathrm{vel}}
=
\left\|
\hat{\dot{\mathbf{q}}}_{t:t+H}
-
\dot{\mathbf{q}}_{t:t+H}
\right\|_1 ,
\end{equation}
where $\dot{\mathbf{q}}_{t:t+H}$ is estimated from consecutive reached
positions in the demonstrations.
To explicitly couple position and velocity within each action chunk, we further
introduce
\begin{equation}
\mathcal{L}_{\mathrm{con}}
=
\frac{1}{H}
\sum_{h=0}^{H-1}
\left\|
\frac{
\hat{\mathbf{q}}_{t+h+1}
-
\hat{\mathbf{q}}_{t+h}
}{
\Delta t
}
-
\hat{\dot{\mathbf{q}}}_{t+h}
\right\|_1 .
\end{equation}
This constraint aligns the motion implied by the predicted position sequence
with its velocity representation, explicitly regularizing intra-chunk motion
evolution.
Finally, the CVAE latent space is regularized by
\begin{equation}
\mathcal{L}_{\mathrm{KL}}
=
D_{\mathrm{KL}}
\left(
Q(\mathbf{z}\mid\mathbf{q}_t,\mathbf{q}_{t:t+H})
\parallel
\mathcal{N}(\mathbf{0},\mathbf{I})
\right).
\end{equation}

\section{Experiments}
\begin{figure}[!t]
    \centering
    \includegraphics[width=\linewidth]{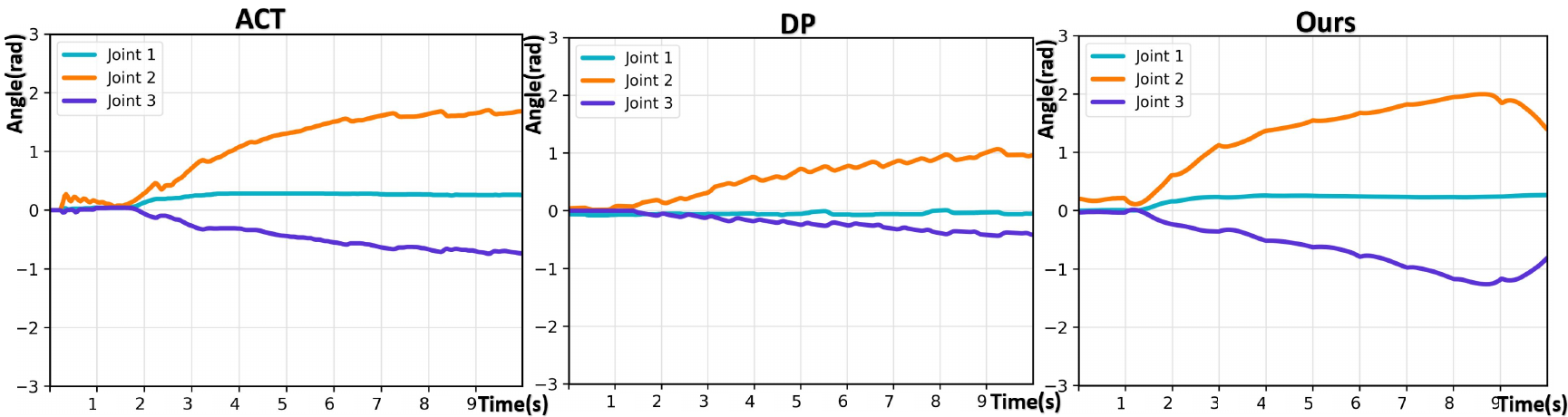}
    \vskip-1ex
    \captionsetup{justification=justified,singlelinecheck=false}
    \caption{Joint-angle trajectories of Piper Joints 1--3 during the Close Drawer task. Ours yields smoother trajectories with reduced oscillation than ACT~\cite{Zhao2023ACT}and DP~\cite{Chi2023DiffusionPolicy}.}
    \label{fig:8}
    \vskip-3ex
\end{figure}

\subsection{Experiment Setup}\label{experiment setup}
\textbf{Data Preparation and Collection:} 
We decouple articulated-object mobile manipulation into task-oriented navigation and manipulation and prepare the corresponding inputs and training data accordingly.
As shown in Fig.~\ref{fig:10}, the mobile manipulation setup uses a Unitree Go2-W and two Piper arms.
For navigation, we design structured prompts based on PP-CoT (see Sec.~\ref{subsec:COT}) to guide the MLLM in a zero-shot manner, enabling high-level manipulation instructions to be decomposed into sequential low-level navigation subgoals for coarse target-directed navigation.
After reaching the target vicinity, DINOv2-based visual features, a pretrained diffusion policy, and Kalman filtering are used for fine-grained approach adjustment until a manipulation-ready base configuration is reached.
For manipulation, we collect high-quality expert demonstrations through low-latency master-follower teleoperation, where joint-velocity references are incorporated into impedance-based tracking with $k_p=8$ and $k_d=0.1$ to preserve both position and velocity information in expert motions. 
The master-follower control loop updates joint-position and joint-velocity references at $220\,\mathrm{Hz}$, while the reached follower joint positions, finite-difference velocities, and camera observations are recorded at $30\,\mathrm{Hz}$.
To fairly evaluate the effectiveness of exploiting velocity information through position-velocity coupling, both variants use the same joint-space teleoperation framework and experimental conditions. During data collection, only joint-position commands are transmitted. During inference, we set $k_d=0$ to remove the velocity term and obtain a position-only baseline.

\textbf{Implementation Details:}
As shown in Fig.~\ref{fig:4}, for navigation, we implement PP-CoT as a zero-shot, two-stage inference framework using Qwen3.7-Max, where perception and planning are executed sequentially with the perception output passed to the planning stage. Both stages use a thinking budget of $2{,}048$ tokens, a sampling temperature of $0.5$, a top-$k$ value of $20$, and a fixed random seed of $1234$, while the maximum answer length and request timeout are set to $256$ tokens and $120$ seconds per stage, respectively. For manipulation, the model uses an ImageNet-pretrained ResNet-18 and an ACT-based CVAE trained with AdamW for $100{,}000$ steps, with a learning rate of $0.00001$, a weight decay of $0.0001$, and a batch size of $8$. The Transformer has a hidden dimension of $512$, eight attention heads, four encoder layers, one decoder layer, a $32$-dimensional latent space, and an action chunk size of $30$, with an auxiliary velocity head predicting joint velocities alongside the position action head. Experiments are conducted on two NVIDIA RTX 3090 GPUs, and the objective combines L1 position loss, velocity supervision weighted by $0.1$, motion consistency weighted by $0.1$ with $\Delta t=1/30\,\mathrm{s}$, and KL regularization weighted by $10$.
\begin{figure}[!t]
    \centering
    \includegraphics[width=\linewidth]{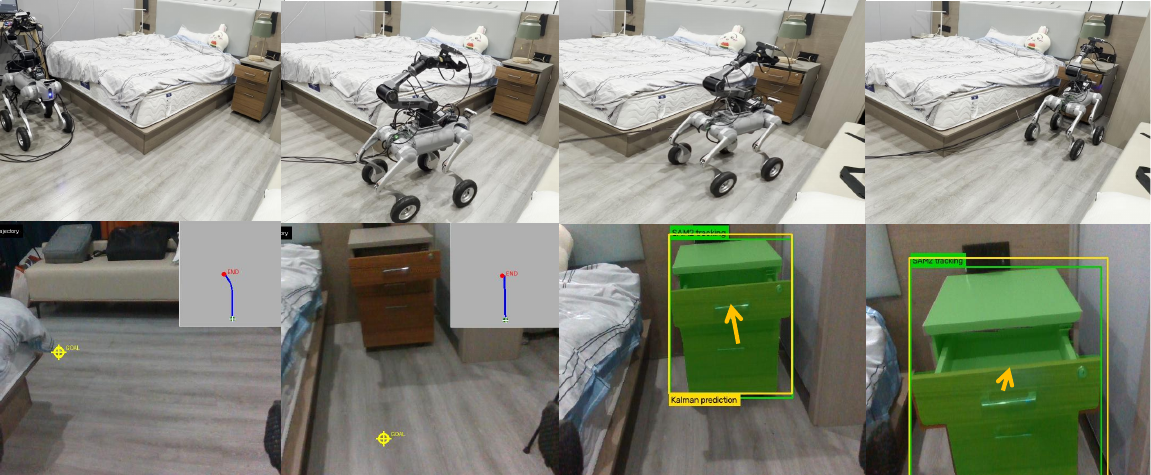}
    %\vskip-1ex
    \captionsetup{justification=justified,singlelinecheck=false}
    \caption{Real-world demonstration of AAM. The first two frames show coarse approach, and the last two refine base orientation and target distance for manipulation readiness.}
    \label{fig:9}
    %\vspace{-17pt}
    \vskip-2ex
\end{figure}

\subsection{Real-World Experimental Comparison}\label{Baseline Comparison}

\textbf{Quantitative Results:}
We first evaluate task-oriented navigation across three articulated-object mobile manipulation tasks. 
As shown in Table~\ref{tab:1}, TONAV achieves success rates of $80.00\%$, $60.00\%$, and $40.00\%$ on Close Drawer, Turn On Lamp, and Lower Toilet Lid, respectively, yielding an overall SR of $60.00\%$. 
This outperforms StreamVLN~\cite{DBLP:conf/icra/WeiWYMCZCYWCWL26} ($13.33\%$) and InternNav~\cite{wang2025internvla} ($26.67\%$) by absolute margins of $46.67$ and $33.33$ percentage points, respectively.

To further analyze the contribution of PP-CoT (see Sec.~\ref{subsec:COT}) and the effect of different LLMs, we conduct additional ablation experiments in Table~\ref{tab:3}. 
Removing PP-CoT reduces the overall SR of Doubao-Seed-2.1-Pro and Qwen-3.7-Max to $13.33\%$ and $26.67\%$, respectively, whereas incorporating PP-CoT improves their performance to $46.67\%$ and $60.00\%$. 
Moreover, Qwen-3.7-Max consistently outperforms Doubao-Seed-2.1-Pro, demonstrating stronger capability in decomposing high-level manipulation instructions into executable navigation subgoals.

The navigation improvement further benefits from the progressive navigation design. 
PP-CoT grounds task-relevant targets and landmarks and decomposes high-level instructions into executable navigation subgoals for coarse target-directed navigation. 
As shown in Fig.~\ref{fig:9}, the robot first progressively approaches the target region in a coarse-grained manner (the first two frames). 
After reaching the target vicinity, AAM (see Sec.~\ref{sec:AAM}) takes over local base control and further adjusts the relative orientation and distance between the robot and the target through target tracking, delayed-observation compensation, and alignment-distance regulation (the last two frames), ultimately reaching a manipulation-ready configuration. 
This progressive process effectively reduces the gap between coarse navigation termination and the base configuration required for downstream manipulation.

For manipulation, all methods start from the same manipulation-ready configurations for a fair comparison. 
As shown in Table~\ref{tab:2}, without position-velocity-coupled teleoperation, ACT~\cite{Zhao2023ACT} and DP~\cite{Chi2023DiffusionPolicy} achieve overall success rates of $13.33\%$, while TONAV reaches $53.33\%$. 
With position-velocity expert demonstrations, TONAV further improves the overall success rate to $80.00\%$, outperforming ACT~\cite{Zhao2023ACT} and DP~\cite{Chi2023DiffusionPolicy} by absolute margins of $33.33$ and $53.33$ percentage points, respectively, while achieving $80.00\%$ success across all three tasks. 
The improvement mainly comes from two aspects. 
First, position-velocity-coupled teleoperation reduces the motion mismatch between the master and follower arms and provides higher-quality demonstrations with more consistent position-velocity trajectories. 
Second, the proposed learning objective explicitly exploits these demonstrations: $\mathcal{L}_{\mathrm{vel}}$ preserves the expert motion dynamics, while $\mathcal{L}_{\mathrm{con}}$ couples predicted positions and velocities to regularize intra-chunk motion evolution. 
Together, high-quality position-velocity demonstrations and explicit motion-consistency modeling reduce motion discontinuities and improve stable contact along constrained articulated-object trajectories.

\begin{table}[!t]
\centering
\caption{
Comparison of complete mobile manipulation success rates, with all methods sharing a single navigation run to the manipulation-ready region. (\textit{w/o P-V Control}: demonstrations collected without position-velocity-coupled teleoperation.)}
\label{tab:2}
\resizebox{\columnwidth}{!}{
\begin{tabular}{lcccc}
\toprule
& \multicolumn{3}{c}{\textbf{Task Success Rates}}
& \textbf{Overall} \\
\cmidrule(lr){2-4}
\textbf{Method}
& \textbf{Close Drawer}
& \textbf{Turn On Lamp}
& \textbf{Lower Toilet Lid}
& \textbf{SR} \\
\midrule

ACT~\cite{Zhao2023ACT} (w/o P-V Control)
& 1/5 & 0/5 & 1/5 & 13.33\% \\

DP~\cite{Chi2023DiffusionPolicy} (w/o P-V Control)
& 1/5 & 1/5 & 0/5 & 13.33\% \\

TONAV (w/o P-V Control)
& 2/5 & 2/5 & 4/5 & 53.33\% \\

\midrule

ACT~\cite{Zhao2023ACT}
& 2/5 & 3/5 & 2/5 & 46.67\% \\

DP~\cite{Chi2023DiffusionPolicy}
& 2/5 & 1/5 & 1/5 & 26.67\% \\

\textbf{TONAV (Ours)}
& \textbf{4/5}
& \textbf{4/5}
& \textbf{4/5}
& \textbf{80.00\%} \\

\bottomrule
\end{tabular}
}
\vskip-2ex
\end{table}

\textbf{Qualitative Analysis:}
We first analyze the navigation behaviors in Fig.~\ref{fig:7}. 
Although StreamVLN~\cite{DBLP:conf/icra/WeiWYMCZCYWCWL26} and InternNav~\cite{wang2025internvla} can follow the decomposed subgoals and reach the target vicinity, their terminal base poses are insufficiently conditioned on the downstream manipulation task.
As a result, the manipulator may approach or reach its kinematic limits before establishing effective interaction with the target object, making subsequent manipulation difficult or even infeasible.

In contrast, TONAV combines PP-CoT (see Sec.~\ref{subsec:COT}) with AAM (see Sec.~\ref{sec:AAM}) to progressively refine the robot pose toward a manipulation-ready configuration, providing a more suitable base position and orientation for downstream interaction.
\begin{table}[!t]
\centering
\caption{
Ablation of PP-CoT and comparison of different LLMs for manipulation-oriented navigation across three articulated-object mobile manipulation tasks.
}
\label{tab:3}
\resizebox{\columnwidth}{!}{
\begin{tabular}{lcccc}
\toprule
& \multicolumn{3}{c}{\textbf{Instruction Decomposition Success Rates}} 
& \textbf{Overall} \\
\cmidrule(lr){2-4}
\textbf{Method}
& \textbf{Close Drawer} 
& \textbf{Turn On Lamp}
& \textbf{Lower Toilet Lid}
& \textbf{SR}\\
\midrule

Doubao-Seed-2.1-Pro (w/o PP-CoT)
& 1/5 & 1/5 & 0/5 & 13.33\% \\

Qwen-3.7-Max (w/o PP-CoT)
& 2/5 & 1/5 & 1/5 & 26.67\% \\

\midrule

Doubao-Seed-2.1-Pro
& 3/5 & 2/5 & 2/5 & 46.67\% \\

\textbf{Qwen-3.7-Max (Ours)}
& \textbf{4/5}
& \textbf{3/5}
& \textbf{2/5}
& \textbf{60\%} \\

\bottomrule
\end{tabular}
}
\vskip-2ex
\end{table}
We then compare the manipulation behaviors. 
As shown in Fig.~\ref{fig:6}, TONAV performs more stable and complete interactions on both lamp switching and toilet-lid lowering. Compared with ACT~\cite{Zhao2023ACT} and DP~\cite{Chi2023DiffusionPolicy}, our method maintains more reliable end-effector contact while following the constrained object motion, resulting in fewer contact losses and more continuous task completion.

The joint trajectories in Fig.~\ref{fig:8} further show that TONAV exhibits less oscillation and smoother motion evolution. This improvement mainly benefits from velocity supervision and the position-velocity consistency constraint, which preserve motion dynamics and reduce discontinuities across action chunks,\begin{wrapfigure}{r}{0.21\textwidth}
    \centering
    \includegraphics[width=0.21\textwidth]{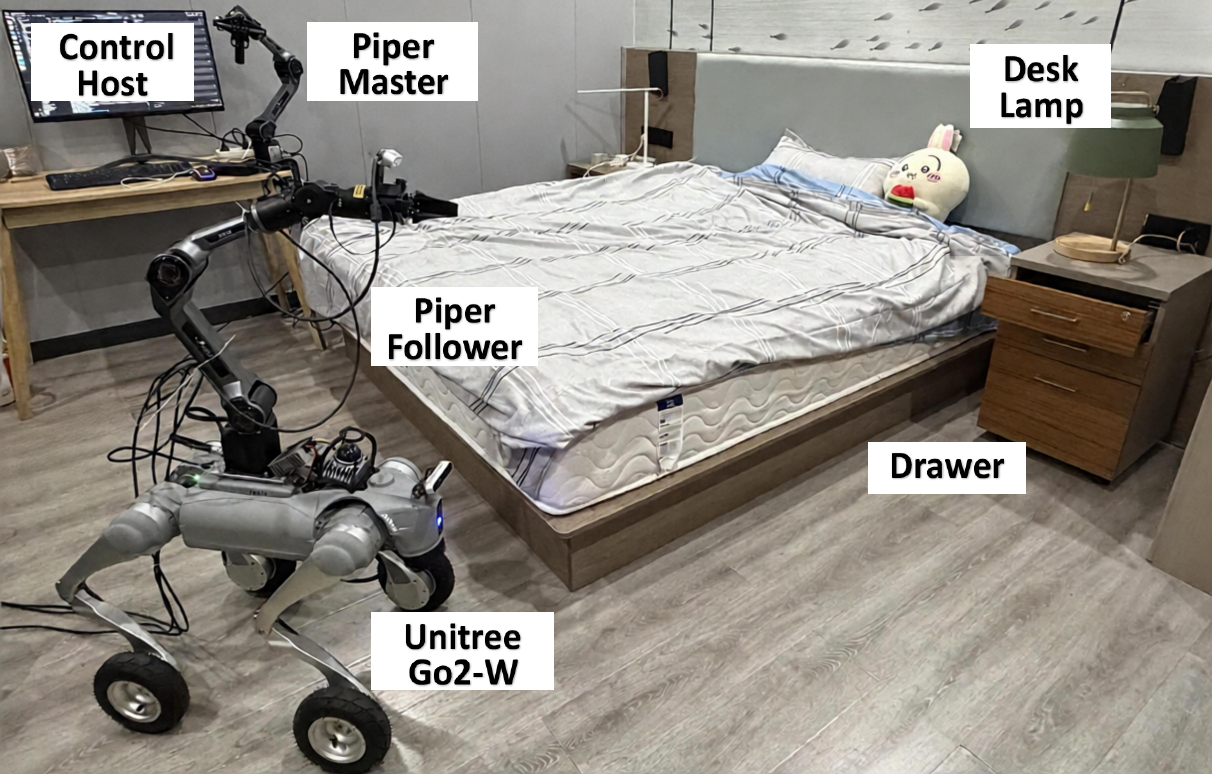}
    \vskip-1ex
    \caption{Real-world experiment setting.}
    \vskip-3ex
    \label{fig:10}
\end{wrapfigure} thereby improving sustained-contact stability. Moreover, temporal inconsistency in expert demonstrations can propagate to the learned policy and amplify tracking errors during contact-rich interaction. Our position-velocity-coupled teleoperation (see Sec.~\ref{subsec:teleoperation}) reduces master-follower tracking lag and improves temporal consistency, as shown in Fig.~\ref{fig:2}, providing higher-quality demonstrations for policy learning.

\section{Conclusion}
This work presents a unified framework for articulated-object quadrupedal mobile manipulation, combining progressive task-oriented navigation with position-velocity-coupled teleoperation and action-velocity chunk learning. 
It enables robots to reach manipulation-ready configurations and perform smooth, stable sustained-contact manipulation. Real-world experiments demonstrate improved navigation and manipulation performance with reduced motion oscillation and contact instability. Future work will explore tactile sensing and richer multimodal feedback to improve robustness against contact disturbances in long-horizon sustained-contact tasks and generalization across varying scene layouts and initial configurations.

{\small
\bibliographystyle{IEEEtran}
\bibliography{references}
}

\end{document}